\documentclass[default,iicol]{sn-jnl}

\usepackage{graphicx}%
\usepackage{amsmath,amssymb,amsfonts}%
\usepackage{amsthm}%
\usepackage{mathrsfs}%
\usepackage[title]{appendix}%
\usepackage{xcolor}%
\usepackage{textcomp}%
\usepackage{manyfoot}%
\usepackage{booktabs}%
\usepackage{booktabs}\let\cline\cmidrule
\usepackage{algorithm}%
\usepackage{algorithmicx}%
\usepackage{algpseudocode}%
\usepackage{listings}%
\usepackage{makecell,multirow,diagbox, multicol}
\usepackage[switch]{lineno}
\usepackage{etoolbox}

\theoremstyle{thmstyleone}%
\theoremstyle{thmstyletwo}%

\theoremstyle{thmstylethree}%

\begin{document}
	
	\title[Article Title]{Machine Learning Modeling for Multi-order Human Visual Motion Processing}

	\author{\fnm{Zitang} \sur{Sun}}\email{sun.zitang.c09@kyoto-u.jp}
	\author{\fnm{Yen-Ju} \sur{Chen}}\email{chen.yen-ju.t05@kyoto-u.jp}
	\author{\fnm{Yung-Hao} \sur{Yang}}\email{yang.yunghao.8v@kyoto-u.ac.jp}
	\author{\fnm{Yuan} \sur{Li}}\email{li.yuan.67n@st.kyoto-u.ac.jp}
	\author*{\fnm{Shin'ya} \sur{Nishida*}}\email{nishida.shinya.2x@kyoto-u.ac.jp}
	
	\affil{\orgdiv{Graduate School of Informatics}, \orgname{Kyoto University}, \orgaddress{ \city{Kyoto}, \postcode{606-8501},  \country{Japan}}}

	
	\abstract{	
		Our research aims to develop machines that learn to perceive visual motion as do humans. While recent advances in computer vision (CV) have enabled DNN-based models to accurately estimate optical flow in naturalistic images, a significant disparity remains between CV models and the biological visual system in both architecture and behavior. This disparity includes humans' ability to perceive the motion of higher-order image features (second-order motion), which many CV models fail to capture because of their reliance on the intensity conservation law. Our model architecture mimics the cortical V1-MT motion processing pathway, utilizing a trainable motion energy sensor bank and a recurrent graph network. Supervised learning employing diverse naturalistic videos allows the model to replicate psychophysical and physiological findings about first-order (luminance-based) motion perception. For second-order motion, inspired by neuroscientific findings, the model includes an additional sensing pathway with nonlinear preprocessing before motion energy sensing, implemented using a simple multilayer 3D CNN block. When exploring how the brain acquired the ability to perceive second-order motion in natural environments, in which pure second-order signals are rare, we hypothesized that second-order mechanisms were critical when estimating robust object motion amidst optical fluctuations, such as highlights on glossy surfaces. We trained our dual-pathway model on novel motion datasets with varying material properties of moving objects. We found that training to estimate object motion from non-Lambertian materials naturally endowed the model with the capacity to perceive second-order motion, as can humans. The resulting model effectively aligns with biological systems while generalizing to both first- and second-order motion phenomena in natural scenes.
	}


	\maketitle
	\section{Introduction}\label{sec1}

		Creating machines that perceive the world as humans do poses a significant interdisciplinary challenge bridging cognitive science and engineering. From the former perspective, developing human-aligned computational models advances our understanding of brain functions and the mechanisms underlying perception \cite{yamins2016using,kriegeskorte2015deep,wichmann2023deep}. On the latter side, such models, which accurately simulate human perception in diverse real-world scenarios, would enhance the reliability and utility of human-centered technologies.

		Recent advances in deep learning have rendered machine vision comparable to or better than that of humans when performing many vision tasks \cite{deng2009imagenet,fcn}. Visual motion estimation \cite{flownet} is not an exception. State-of-the-art (SOTA) computer vision (CV) models are more accurate than humans in terms of estimating Lagrangian motion map (i.e. optical flow) in natural images \cite{Yang2023}. However, such models are not yet sufficiently human-aligned, being unable to predict human perception in many aspects. For example, CV models are often unstable under certain experimental conditions \cite{sun2023comparative, ranjan2019attacking}. They do not reproduce human visual illusions or fully capture biases inherent in human perception \cite{Yang2023}. Several gaps raise the fundamental question: What is the relationship between DNN-based models and the brain in visual motion processing, and to what extent can DNNs replicate brain functions? 
		
		Recent attempts to integrate insights from cognitive science with deep learning techniques (e.g., MotionNet \cite{rideaux2020but}, DorsalNet \cite{mineault2021your}) demonstrate the DNNs' potential to align with the biological visual motion processing, but these models can not replicate human-like capability to accurately infer the object motion, unlike CV models. Here, we seek not only to define the correspondence between existing DNNs and neural systems but also to develop a model that balances biological plausibility with human-like perceptual responses across broad aspects of motion phenomena and illusions. From an engineering standpoint, we strive to create a human-aligned motion processing model while maintaining motion estimation capabilities comparable to human and SOTA CV models.
		
		In our design, the model features a two-stage process that simulates the cortical V1-MT processing system of primates \cite{simoncelli1998model, nishimoto2011three}. The first stage mimics the function of the primary visual cortex (V1), featuring neurons with multiscale spatiotemporal filters that extract local motion energy (ME). Unlike previous V1 models, the filter tunings are learnable to fit natural optic flow computation. The second stage, which mimics the middle temporal cortex (MT), addresses motion integration and segregation. We introduce a graph network to construct a motion graph of a dynamic scene, enabling flexible connections between local motion elements and recurrent global motion integration. Efficient motion integration aside, the motion graph implicitly encodes object interconnections in a graph topology. Training-free graph cuts\cite{shi2000normalized} can be seamlessly applied for object-level segmentation.
		
		The early version of our model, reported in part in \cite{sun2023modeling}, featured a single-channel motion sensing pathway in the first stage and was trained to estimate the ground truth (GT) flow across various video datasets. Although the model successfully replicated a wide range of findings on biological visual motion processing for low-level, luminance-based motion (first-order motion), it was limited by its motion energy sensors. This limitation rendered it theoretically incapable of explaining higher-level human motion perception involving spatiotemporal pattern processing, such as second-order motion \cite{secondordermotion,cavanagh1989motion}.
		
		Second-order motion, also termed non-Fourier motion, features high-level spatiotemporal features, including spatial or temporal contrast modulations. Such motion perception is observed across many species, including macaques \cite{secmacaque}, flies \cite{secflies}, and humans \cite{baker1999central,cavanagh1989motion}, yet it remains undetectable by most CV models \cite{sun2023comparative}. This limitation stems from CV models' reliance on flow estimation algorithms based on the intensity conservation law \cite{fleet2006optical}, which estimates pixel shifts by matching intensity distributions before and after the movement.

		\begin{figure*}[!h]
			\begin{center}
				\includegraphics[width=1.0\textwidth]{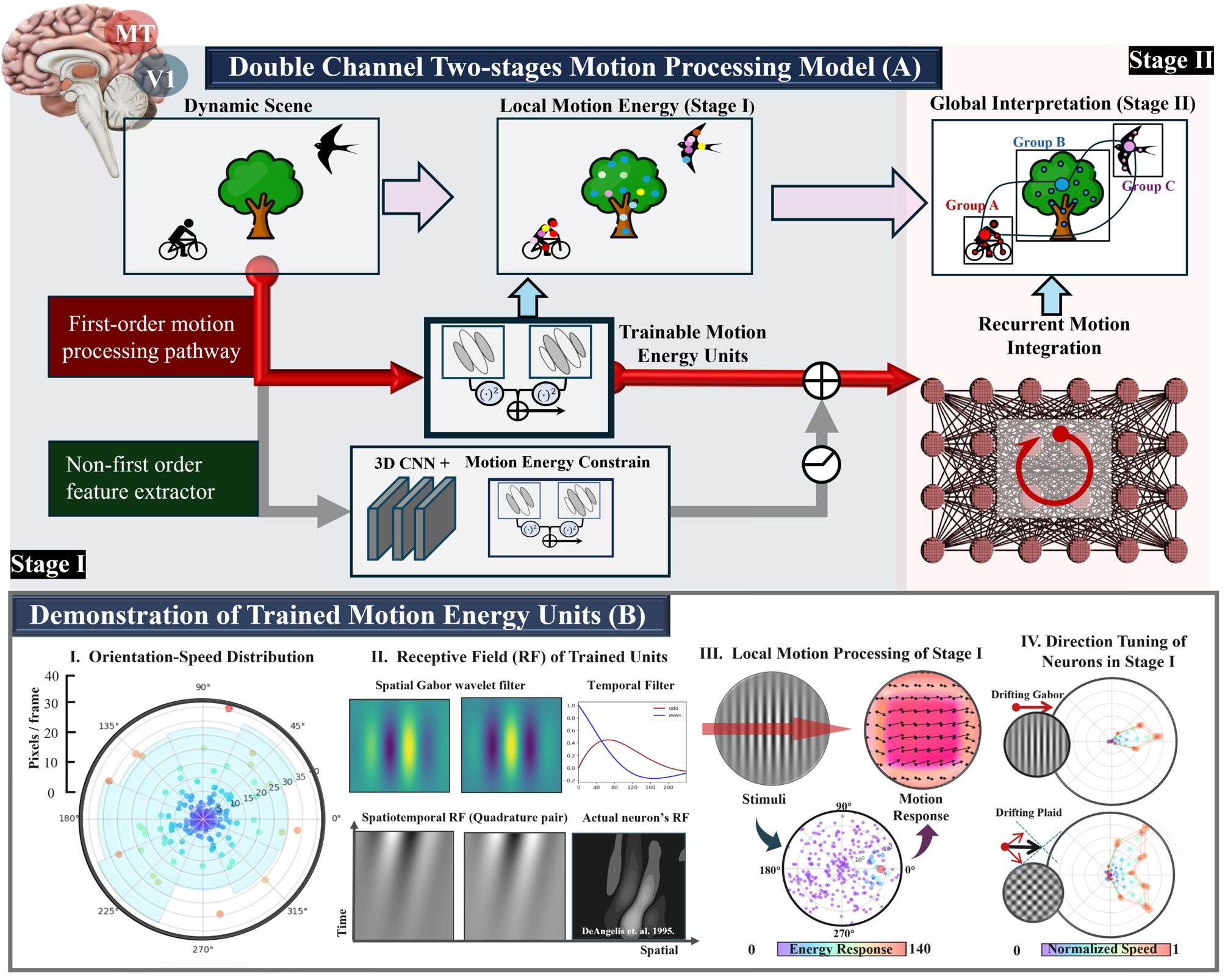}
			\end{center}
			\caption{ {\bfseries \textit{The Two-Stage Motion Perception System Prototype}} {\bfseries (A)}: The first stage (Stage-I) mimics V1 function by detecting local motion. The second stage (Stage II) leverages a graph network for recurrent motion integration and segregation. Stage I incorporates dual channels to process both first-order and higher-order motion. The first-order channel captures Fourier-based motion in motion energy units, thus via the red route. The higher-order channel employs normative 3D CNN layers to extract higher-order features via the gray route. Natural videos are used to train the entire model to estimate motion flow. {\bfseries (B)}: Illustration of spatiotemporal receptive fields of motion energy units in Stage I, resembling those of V1 neurons, which are sensitive to drifting gratings with specific spatiotemporal frequencies and directions of movements.}
			\label{fmodel}
		\end{figure*} 
		
		To encompass both first- and second-order motion perception, we revised the model's structure and training scheme. Inspired by earlier studies on human vision suggesting separate processing mechanisms for first- and second-order motion \cite{clifford1998first,ledgeway1994evidence,smith1998processing}, we introduced a secondary sensing pathway with a naive 3D CNN preceding the motion energy sensing stage \cite{clifford1998first,nishida2000hierarchical}. The 3D CNN is designed to perform nonlinear preprocessing to extract spatiotemporal textures, following the filter-rectify-filter model of second-order motion processing \cite{prins2003mechanism}. Given the computational power of neural networks, the modified model was expected to detect second-order motion after training on an adequate number of artificial, second-order motion stimuli. However, such training is unrealistic in natural environments, where pure second-order motions are rarely observed. The critical scientific question is how and why the biological visual system naturally acquires the ability to perceive second-order motion. 
		
		We hypothesized that second-order motion perception aided the estimation of the motion of objects exhibiting different material properties. Natural non-Lambertian optical effects, such as specular reflections and transparent refractions, can significantly alter the light path of an object. This generates complex and dynamic optical turbulence on the surface of the moving object. Such optical turbulence means that models capturing only first-order motion do not, in fact, accurately estimate object motion. However,  detecting second-order motion, including the movement of dynamic luminance noise, may facilitate the inference of global object motion, particularly when optical fluctuations develop in natural scenes. As a proof of concept that detecting second-order motion correlates with estimating non-Lambertian objects' motion, we created two versions of a novel motion dataset. One contained purely Lambertian (matte) objects, and the other non-Lambertian objects experienced optical turbulence imparted by specular reflections and transparent refractions. We trained different models on both datasets and found that, given an appropriate structure and training environment, the model naturally developed the ability to perceive second-order motion comparable to human capabilities. Finally, we show that our human-aligned visual motion model, with the ability to process both first- and second-order motions, can robustly estimate object motion under noisy natural environments.

		The contributions of our study can be summarized as follows:
		\begin{itemize}
			\item By simulating biological computation using a deep learning framework, we modeled human visual motion processing using trainable motion energy sensing and a graph network for global scene parsing. The dual-channel design further enables the inference of both first-order and second-order motion.
			
			\item The model replicates psychophysical and physiological findings from human and animal studies, achieves dense optical flow estimation comparable to SOTA CV models, and can be extended to segmentation tasks without additional training.
			
			\item Using learning-based modeling, we tested and verified a new scientific hypothesis: Animals may have evolved the capacity to perceive second-order motion because this aids reliable estimation of object-level movement even in the presence of the optical noise of a non-Lambertian optic environment.
		\end{itemize}

	\section{Results}

		In Section \ref{sec2.1}, we present the processing pipeline of the dual-channel motion model. There are both local and global motion processing stages. Section \ref{sec2.2} demonstrates how the model integrates motions. The tasks ranged from those of vision science experiments to complex natural scenarios. Sections \ref{sec2.3} and \ref{sec2.4} extend to higher-order processing,  exploring the relationship between material properties and the ability of second-order motion perception. We used defined benchmarks to quantify the contributions made by first- and higher-order motion processing and compared these to those of representative machine vision models.\footnote{A demo of our project is available on the \url{https://anoymized.github.io/motion-model-website/}. We recommend viewing the dynamic results for a more insightful experience.}
		
		\begin{figure*}[h!]
			\begin{center}
				\includegraphics[width=1.0\textwidth]{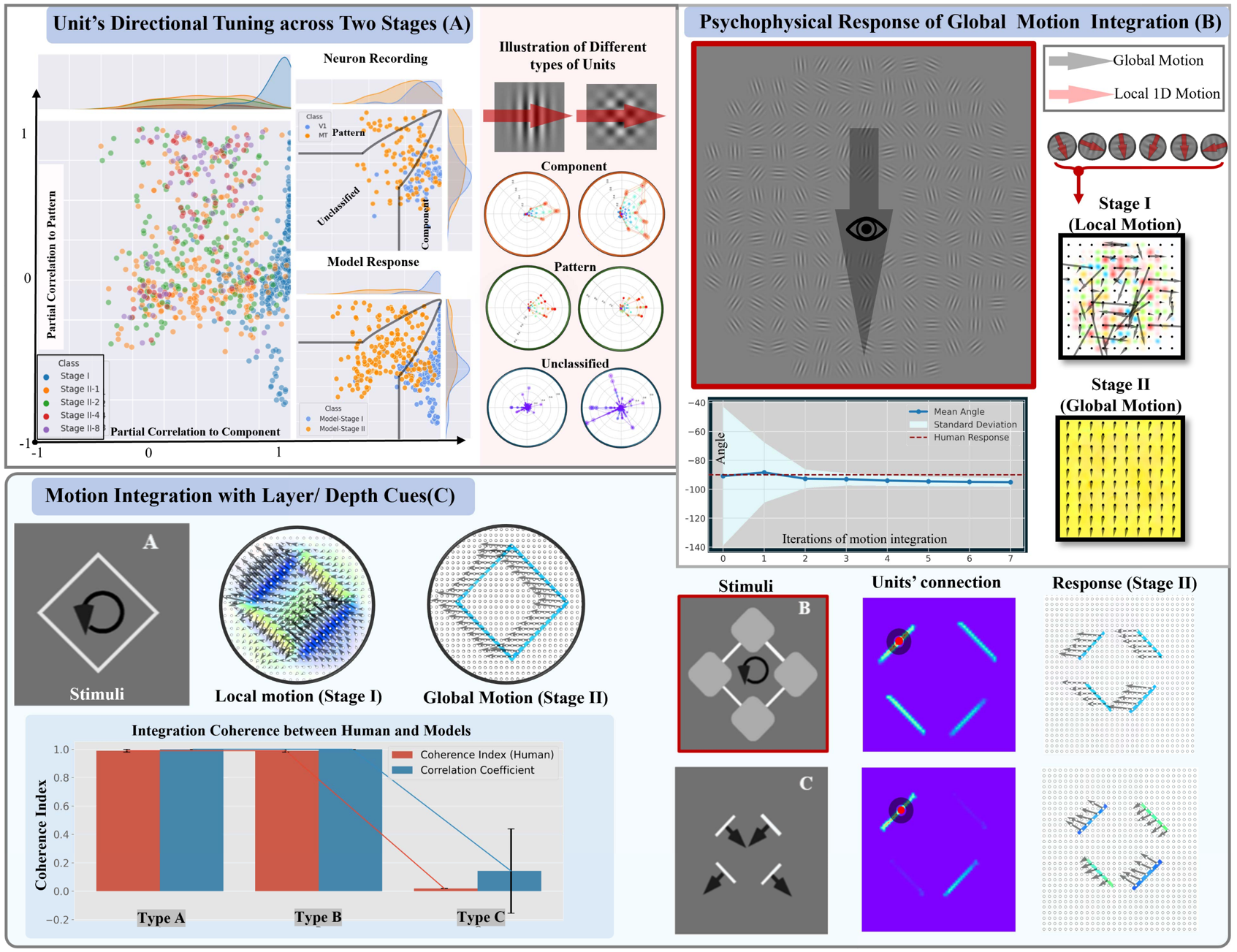}
			\end{center}
			\caption{ {\bfseries \textit{Recurrent Motion Integration and Evaluation in Experimental Setting.}} {\bfseries (A):}   Direction tuning across the neuronal groups when both 1D and 2D stimuli were imparted. The animal data are redrawn from \cite{movshon1992analysis}.  	{\bfseries (B):} Motion integration connects local movements to solve the aperture problem and to globally interpret motion. This aligns with human perception of adaptive pooling \cite{amano2009adaptive}.
				{\bfseries (C)} Motion integration is sensitive to higher-order pattern cues. We used the three scenarios A, B, and C detailed in \cite{mcdermott2001beyond}. The extents of integration were quantified by correlating the directions of motion between adjacent segments across a single rotation cycle. As compared to scenario C, scenario B, characterized by structural constraints and depth cues, led to an increased integration index in the model, similar to human perception \cite{mcdermott2001beyond}. In the middle column of the right panels
				(for unit connections), we visualize the attention heat map derived from the motion graph, showing the connectivity of the unit (marked by a circle) with other units in Stage II.}
			
			\label{f-motionintegration}
		\end{figure*}
		
		\subsection{The two-stages processing model}
		\label{sec2.1}
		Our prototype model features two-stage motion processing that combines classical ME sensors in Stage I with modern DNNs in Stage II. Stage I captures local ME, simulating the function of the V1 area. In contrast, Stage II focuses on global motion integration and segregation and thus simulates the primary function of the MT area. The red route in Fig. \ref{fmodel}-A is that of classical first-order motion. Specifically, we built 256 trainable ME units, each with a quadrature 2D Gabor spatial filter and a quadrature temporal filter. These captured the spatiotemporal motion energies of input videos within a multiscale wavelet space. The key implementation difference from previous ME models is that we embedded computation in the deep learning framework, i.e., each motion energy neuron's parameters, such as preferred moving speed and direction, are trainable to fit the task. In Fig. \ref{fmodel}-B, we demonstrate the speed-direction distribution and filter receptive field of the trained motion energy neurons. These neurons, activated by stimuli with the preferred spatiotemporal frequency configuration, have their activation patterns decoded into perceptual responses (See Fig. \ref{fmodel}-B-III). Due to the motion energy constraints, the activation patterns of the model show a tendency similar to mammalian neuron recordings in the V1 cortex, such as spatiotemporal receptive fields resembling those of V1 neurons and direction-tuning capabilities in Stage I (Fig. \ref{fmodel}-B-III and Fig. \ref{fmodel}-B-IV).
		Moreover, incorporating ME sensors allows the model to replicate human-aligned perception of various motion illusions that are not captured by conventional CV models, such as reverse phi and missing fundamental illusions \cite{sun2023modeling}.

		Stage II incorporates modern DNNs connected to Stage I. It constructs a fully connected graph on local motion energy, treating each spatial location as a node, with all nodes interconnected. We use a self-attention mechanism to define the topological structure of the graph, by which motions are recurrently integrated to generate interpretations of global motion and address aperture problems (Stage II, Fig. \ref{fmodel}-A). A shared trainable decoder is used to visualize the optical flow fields from Stages I and II. The entire model is trained under supervision to estimate pixel-wise object motions in naturalistic datasets \cite{sintel}. We also used a large motion dataset that we constructed specifically for training.
		
		The ME channel captures only first-order motions. To extract information on higher-order motion, we added an alternative channel, depicted by the gray route in Fig. \ref{fmodel}-A. This channel employs trainable multilayer 3D convolutions prior to the ME computations that extract nonlinear spatiotemporal features. This dual-channel design was inspired by earlier vision studies of separate processing design \cite{clifford1998first,ledgeway1994evidence,smith1998processing}. The Methods contain more technical details (Section \ref{sec4.1} and \ref{sec4.2}).

		\begin{figure*}[ht!]
			\begin{center}
				\includegraphics[width=1.0\textwidth]{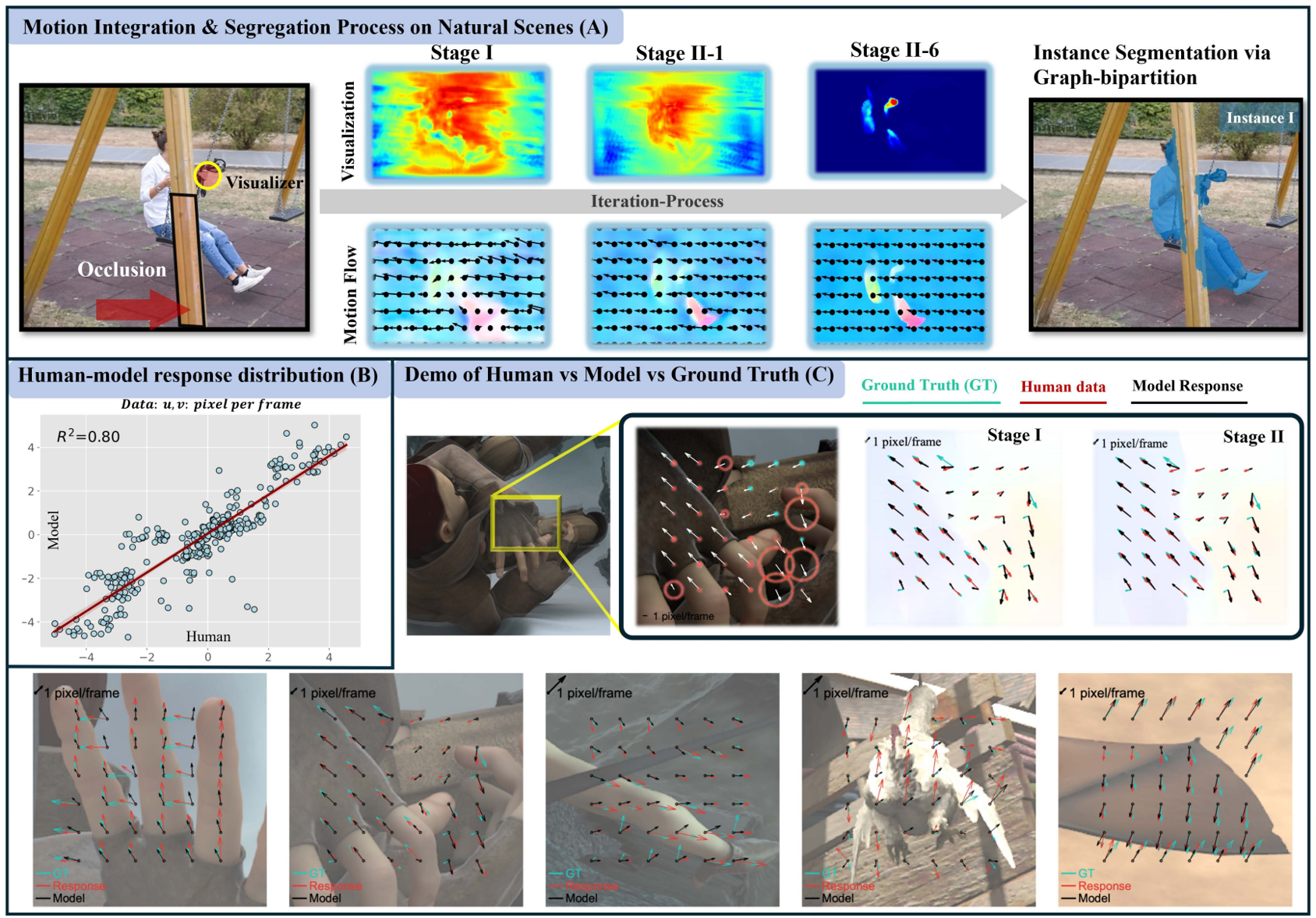}
			\end{center}
			\caption{ {\bfseries \textit{Recurrent Motion Integration and Evaluation in Natural Contexts}}  	{\bfseries (A):} Recurrent integration in natural scenarios. ``Stage II-$N$" refers to the outcomes from the $N$th iteration in Stage II, with visualizations illustrating neuron connectivity via heat maps. The neural connectivity is represented as a graph structure, and through graph bipartitioning \cite{shi2000normalized}, the model can further achieve instance segmentation without any additional training.
				{\bfseries (B, C)}: A comparison of the motion estimations of our model and human perception using the Sintel Benchmark.
				In {\bfseries (B)}, the shadow around the red line represents the 95\% confidence interval (CI) of the linear regression line.
				In {\bfseries (C)}, larger red circles show a closer response to humans over the GT, while blue circles indicate the reverse. Despite being trained to fit the GT, our model uniquely exhibits a higher correlation with human responses than with the GT. } 
			\label{f-motionintegration2}
		\end{figure*}
		
		\subsection{Motion Graph-based Scene Integration}
		\label{sec2.2}
		Stage II of our model emulates the motion integration process, an essential function for linking local motions and solving the aperture problem \cite{apatureproblem}.
		
		Fig. \ref{f-motionintegration}-A (left) displays the responses of 256 units to both drifting-gabor and plaid stimuli. A plaid is two overlapping drifting Gabors \cite{movshon1992analysis}. Analysis revealed three distinct groups of units based on their partial correlations with the Gabor and plaid stimuli. Component cells responded to the direction of a Gabor component. Pattern cells responded to the integrated (coherent) direction of plaid motion. Unclassified cells showed no definitive preference for either response, as shown on the right of Fig. \ref{f-motionintegration}-A. Typically, component cells dominate in the V1 area, while pattern cells, equipped with motion integration capabilities, are more common in the MT and MST regions \cite{movshon1992analysis}. Our model mirrors this biological distribution, as more component cells are in the first stage and more pattern and unclassified cells in the second stage.
		Figure \ref{f-motionintegration}-B shows a global motion of drifting Gabors, where each local patch exhibits a unique local direction and speed but is collectively consistent with unified 2D motion downward. Humans perceive coherent downward motion because they integrate motions across space and orientation, as does our model, Stage I computes local motion. Stage II predicts the direction and speed of global motion, consistent with human psychophysical observations \cite{amano2009adaptive}. In addition, the motion integration process of our model also demonstrates an adaptive pooling of 1D and 2D local motions consistent with human psychophysical data \cite{amano2009adaptive}.
		
		\begin{table*}[h!]
			\caption{{\textbf{\textit{Model v.s. Human v.s. GT. on First-order Motion }}} \boldmath $\rho$ \unboldmath: Partial correlation between human and model while controlling the effects of GT; $r$: Pearson correlation coefficient; $epe $: vector endpoint error; $uv$, $dir$, $spd$ represent motion components in Cartesian space, direction, and speed, respectively.  {\bfseries DorsalNet:} Pre-trained with frozen parameters, with a linear flow decoder to compute dense flow, trained on our dense optical flow datasets. {\bfseries RAFT-val:} the RAFT framework trained on our dataset as a validation. } 
			\label{t1}
			\centering
			\setlength{\tabcolsep}{1.7mm}{
				\begin{tabular}{c|ccc|cccc|cccc} 
					\toprule
					
					\multirow{2}{*}{Method} &\multirow{2}{*}{\boldmath$\rho_{uv}$}& \multirow{2}{*}{\boldmath$\rho_{dir}$} & \multirow{2}{*}{\boldmath $\rho_{spd}$} & \multicolumn{4}{c}{v.s. Human} \vline & \multicolumn{4}{c}{v.s. GT} \\ \cline{5-12}
					&  &   &  &  $  r_{uv}$  & $ {r_{spd}} $ & $  r_{dir}$   & $epe$ &   $r_{uv}$ & $r_{spd}$ & $r_{dir}$ &  $epe$\\ 
					\cline{1-12}
					Farneback \cite{farneback2003two} & 0.27 &0.23&0.11&0.41&0.91&0.34&2.02 &0.34 & 0.33& 0.92&1.96\\
					FlowNet2.0 \cite{flownet} & 0.39& 0.26& 0.34& 0.92& 0.90& 0.96& 0.94&0.95&0.94&0.98& 0.47\\
					RAFT \cite{raft} &0.20&0.22&0.14&0.92&0.90&0.96&0.93&0.98&\bfseries0.99&\bfseries0.99&\bfseries0.25\\
					RAFT-val &0.43& 0.17& 0.42&0.92&0.89&0.96&1.01& 0.92& 0.89& 0.98&0.69\\
					AGFlow \cite{agflow}  &0.30& 0.16& 0.20&0.93&0.90&0.96&0.92& 0.98& 0.98& 0.98&0.27\\
					GMFlow \cite{gmflow} &0.34&0.32&0.17&0.91&0.84&\bfseries0.96&1.03&0.93&0.90&0.97&0.73\\
					FlowFormer \cite{flowformer} &0.36&0.14&0.32&\bfseries 0.93& \bfseries0.91&0.95&\bfseries0.90&\bfseries 0.98 &0.97 &0.98 &0.42\\
					\hline
					FFV1MT \cite{ffv1mt} & 0.31 & 0.16 & 0.31 & 0.83& 0.64 &0.92 & 1.48& 0.59& 0.84&0.94 & 1.29\\
					DorsalNet \cite{mineault2021your} &0.17&0.19&-0.10& 0.20& -0.08&0.86&2.35& 0.20 & -0.04 &0.86 &2.33\\
					\hline
					\bfseries Ours-Stage I & 0.19& 0.17& 0.01&   0.36&0.54&0.84& 2.11 & 0.32 &0.60 &0.84 &2.08 \\
					
					\bfseries Ours-Stage II & \bfseries0.52& \bfseries 0.42& \bfseries 0.45&   0.90&0.88&0.95&0.98&    0.86&0.88&0.96& 0.94\\
					
					\bottomrule
				\end{tabular}
			}
		\end{table*}
		
		Fig. \ref{f-motionintegration}-C illustrates how the model adapts to spatial patterns when integrating motions. When a diamond moves along a circular path (Scenario A), Stage II integrates local motions from the four edges into a coherent global motion. The local ME detectors of Stage I cannot do this, as they detect that the segments move orthogonally. See the left side of Fig. \ref{f-motionintegration}-C).
		In Scenario B, the corners of the diamond are occluded by stationary rounded squares. Despite this, the model integrates the local motions of all four line segments into a single coherent motion. The heatmap of the Stage II connections shows that the line segments remain linked. This cannot be simply attributed to a wide integration window from the motion graph because, in Scenario C, where the occluders are invisible, the connections between the line segments are lost during Stage II, and the model generates incoherent motion.
		The model's behavior across Scenarios A-C aligns well with human psychophysical data \cite{mcdermott2001beyond}, as shown by the similarity in the motion coherence index between the model and humans (see bar plot at the bottom-left of Fig. \ref{f-motionintegration}-C). Notably, in Scenario B, the model appears to consider the spatial relationships between occluders and edge segments, mimicking human interpretation of motion integration.
		
		Stage II is essential when processing complex natural scenes, as demonstrated in Fig. \ref{f-motionintegration2}-A. These scenes often exhibit chaotic local motion energies, compounded by challenges such as occlusions and non-textured regions. Addressing these complexities requires long-range and flexible spatial interactions, which are effectively handled by the graph-based, recurrent integration process of Stage II.
		During the iterative process, the model represents local regions as nodes of a graph. The connection weights between locations are captured by the adjacency matrix \( A \in \mathbb{R}^{HW \times HW} \). This matrix is normalized to within the range \( (0, 1) \), where higher values indicate stronger connections. An affinity heatmap of a specific row of the adjacency matrix is shown in Fig. \ref{f-motionintegration2}-A. This reveals how Stage II distinguishes objects from the background and adaptively establishes connections across occlusions. We hypothesized that some of the information required for object-level segmentation was inherently encoded in the topology of the motion-based graph. To test this, we used a training-free visualization method. Specifically, we engaged in graph bipartitioning based on the eigenvector corresponding to the second smallest eigenvalue of the graph Laplacian \cite{shi2000normalized}. As shown in Fig. \ref{f-motionintegration2}-A, this enabled instance segmentation. The results indicated that the model integrated motion representations and object-level recognitions via graph structure, thus effectively segmenting objects even across occlusions.
		
		Such results suggested that our motion-graph-based integration mechanism might unify motion perception and object segmentation in a single framework. These processes are thought to be closely related in the human visual system \cite{handa2018neuronal}. Through a recurrent process, local motion signals become accurately combined in a graph space, yielding clear object-level representation in a coarse-to-fine manner. See section \ref{sec4.13} for the details.

		\begin{figure*}[!ht]
			\vspace{-5mm}
			\begin{center}
				\includegraphics[width=1.0\textwidth]{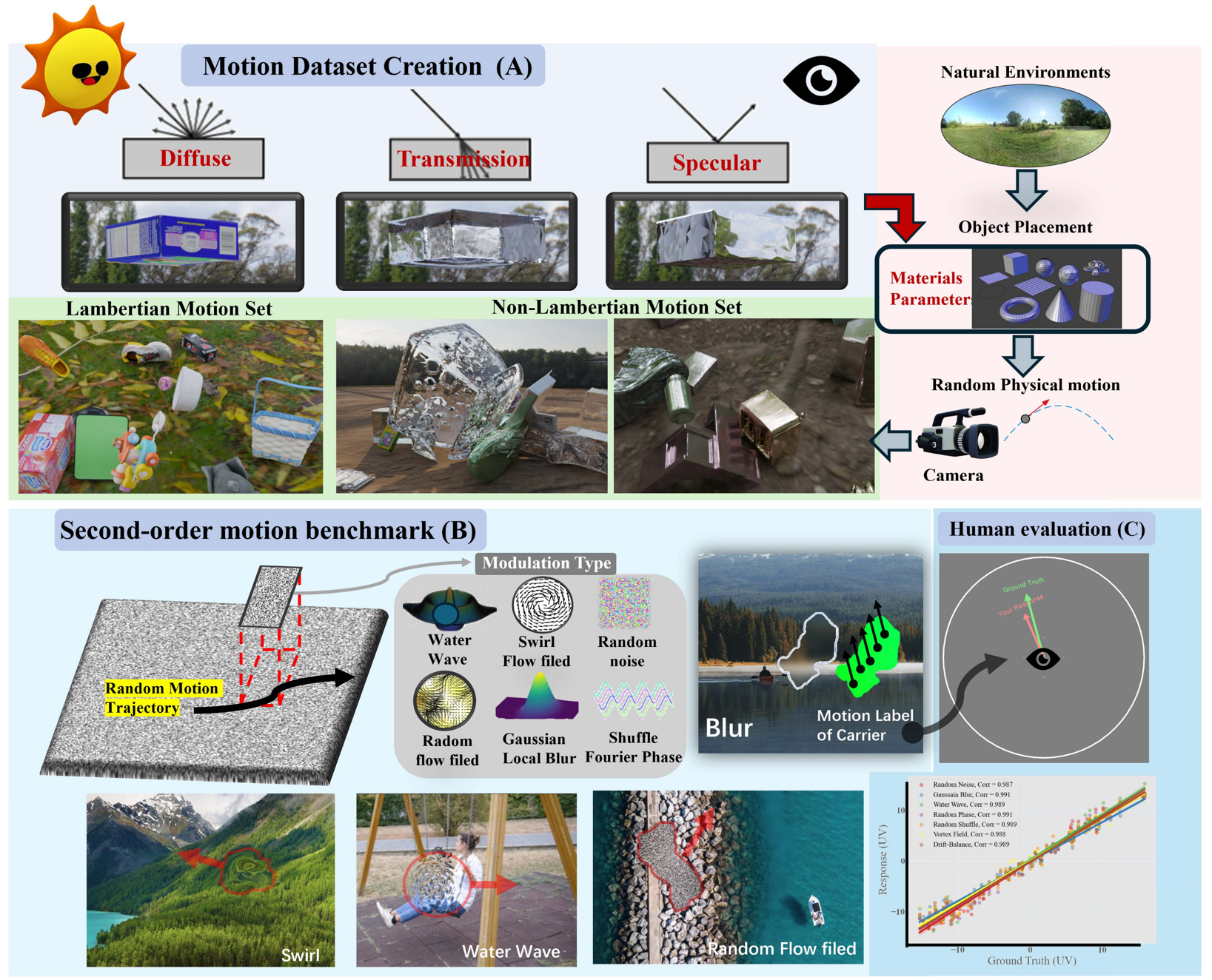}
			\end{center}
			\caption{ {\bfseries \textit{A Material-Controlled Motion Dataset and a Second-order Benchmark Demonstration}} {\bfseries (A):} We manipulated the material properties to create two motion datasets with optical flow labels. One featured purely reflective materials. The other included non-Lambertian materials such as specular, glossy, translucent, and anisotropic surfaces. {\bfseries (B):} A large second-order benchmark was generated by applying naturalistic modulations, thus water waves and swirl effects, to natural images. {\bfseries (C):} The psychophysical experiments performed using these datasets revealed that humans reliably and accurately perceive a range of second-order motions. Machine vision models cannot yet do this. The accompanying figure illustrates the perceived motion vectors from a single participant's data. The shadow around the fitted line represents the 95\% CI} 
			\label{fsecdata}
		\end{figure*}
		
		We further validated the model using the Sintel slow benchmark \cite{sintel}. This naturalistic, high-frame-rate motion dataset incorporates psychophysically measured human-perceived flows \cite{Yang2023}. We compared our model to various CV optical flow estimation methods, including traditional algorithms such as Farneback \cite{farneback2003two}, biologically inspired models \cite{ffv1mt,mineault2021your}, SOTA CV models including multiscale inference methods \cite{flownet,flownet2}, spatial recurrent models \cite{raft}, graph reasoning approaches \cite{agflow}, and vision transformers \cite{gmflow}.
		
		As detailed in Table \ref{t1}, we computed the Pearson correlation coefficients and vector endpoint errors (EPEs) between the model predictions, human responses, and the GT. Crucially, we also derived partial correlations between the human and model responses while controlling the influence of GT. This is essential when evaluating whether a model accurately replicates human responses. A model might appear to well-mimic human responses simply by closely approximating the GT \cite{Yang2023}, which is relatively easy to achieve using optimized CV models. Quantitatively, even though our structure was not specifically optimized in terms of accurate flow prediction, the performance thereof was comparable to those of SOTA CV models when inferring motion in natural scenes. The partial correlations are promising. Fig. \ref{f-motionintegration2}-B illustrates the $(u, v)$ vector distribution of our model in realistic scenarios, showing a strong correlation with human perceptual data. Additionally, Fig. \ref{f-motionintegration2}-C visually demonstrates that motion integration in Stage II introduces biases toward human perception and illusions, deviating from physical GT.

		\begin{figure*}[ht!]
			\begin{center}
				\includegraphics[width=1.0\textwidth]{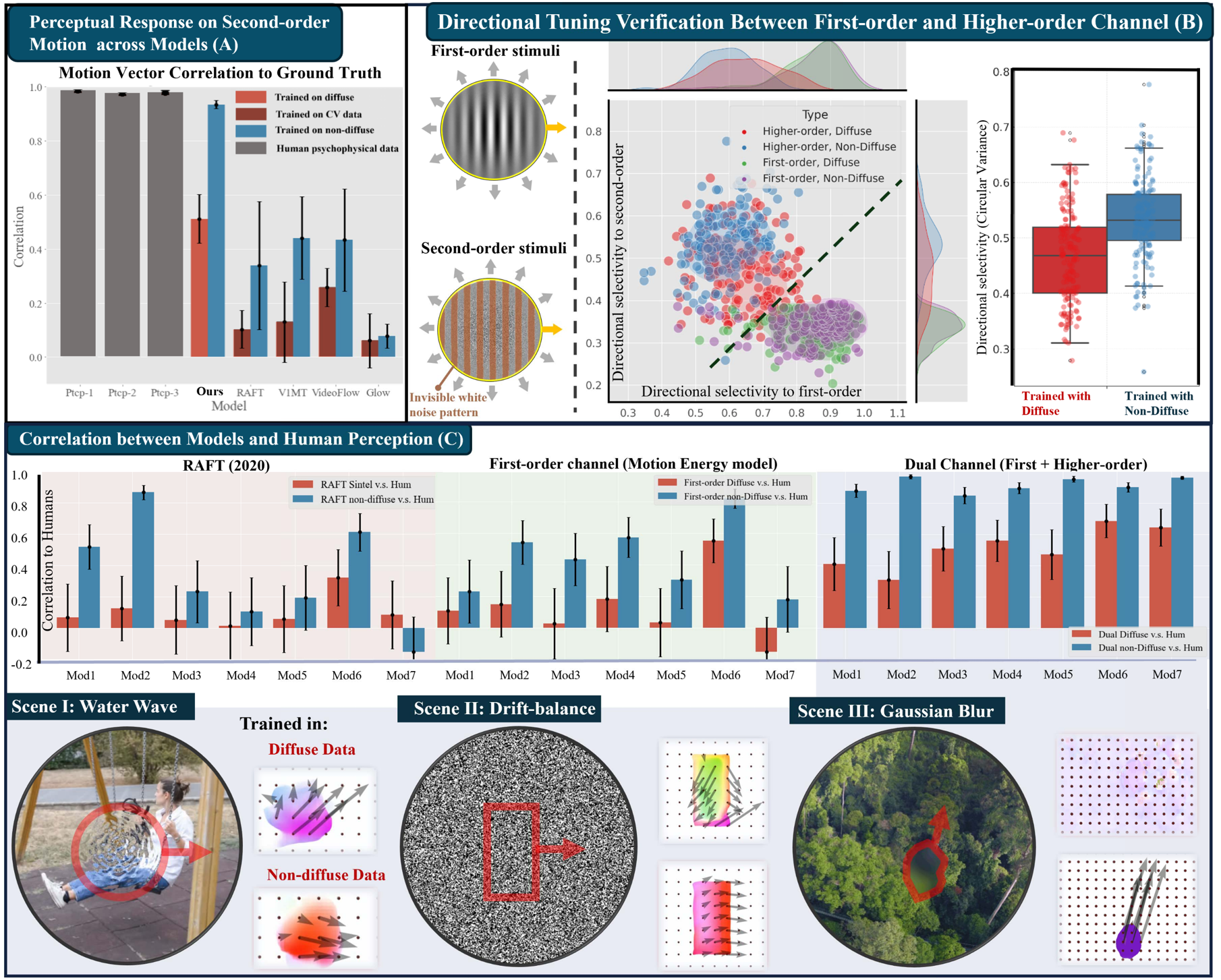}
			\end{center}
			\caption{{\bfseries \textit{The Interplay Between Material Properties and Second-order Motion Perception}}
				{\bfseries (A)} We computed the average Pearson correlations across various types of second-order motion and contrasted these to those of contemporary CV models. Our dual-channel model, trained on a non-diffuse dataset, exhibited superior performance. The proficiency was near-human. The error bars show the 95\% CI across seven modulation types; Ptcp-\textit{N} represent different participants.
				{\bfseries (B)}: The directional tuning curves of model units were tested using both first- and second-order gratings. A modified circular variance measure was employed to quantify directional tuning. This ranged from 0 to 1, where 0 indicates a uniform distribution. We present the results for units in the first- and higher-order channels trained on both the diffuse and non-diffuse datasets. 
				{\bfseries (C)}: Detailed Pearson Correlations with the Human Responses across all Modulation Types. Mod1 to Mod7 are various second-order modulations, including random noise, Gaussian blur, water wave, Fourier phase shuffle, random pixel shuffle, swirl, and drift-balanced motion. The dual-channel model trained on non-diffuse datasets exhibited significantly improved recognition of second-order motion.  The error bars show the 95\% CI within each modulation type.
			}
			
			\label{fig-resultsec}
		\end{figure*}
		\subsection{Material Properties and Second-order Motion Perception}
		\label{sec2.3}
		Despite including a second channel that extracted higher-order features, our model could not identify second-order motion when trained only on existing motion datasets. This limitation reflects broader challenges in CV, as other DNN-based models also fail to capture second-order motion perception in conventional motion estimation tasks \cite{sun2023comparative}. To test our hypothesis that the biological system evolved to perceive second-order motion for estimating object movement amidst optical noise from non-diffuse materials, we constructed datasets that controlled the properties of object materials. One dataset contained ``diffuse'' (matte) reflections and the other ``non-diffuse'' properties, including glossy, transparent, and metallic surfaces (Fig. \ref{fsecdata}-A). The model was trained with a focus on higher-order motion extractors to estimate the GT of object motion while ignoring optical interferences caused by non-diffuse reflections during movement.
		
		To quantify second-order motion perception, we developed a benchmark using natural images with various second-order modulations. As shown in Fig. \ref{fsecdata}-B, the benchmark included classical drift-balanced motion  (temporal contrast modulation) \cite{secondordermotion}; local low contrast (spatial modulation); and natural phenomena such as water waves and swirling flow fields (spatiotemporal modulation). The latter movements are not pure second-order motion but are near-indiscernible in Fourier space, given the chaotic optical disturbances caused by reflection and refraction. Our psychophysical experiment revealed a strong correlation between the physical GT and the human response in terms of detecting second-order motion $(r_{mean}=0.983, sd=0.005)$ (Fig. \ref{fsecdata}-C). In contrast, a representative CV model, RAFT, was associated with a much lower correlation $(r=0.102)$.
		We trained our model on the diffuse and non-diffuse datasets and compared the correlations with human responses. The results of Fig. \ref{fig-resultsec}-C indicate that both the dataset material properties and the model architecture significantly influence the perception of second-order motion. Training of the dual-channel model using non-diffuse materials substantially improved recognition of second-order motion. The average correlation was $0.902$, very close to that of human perceptional data (right side of Fig. \ref{fig-resultsec}-C ). 
		
		Fig. \ref{fig-resultsec}-B shows the direction-tuning capacities of the first- and higher-order motion channels as revealed by in silico-neurophysiological methods. First- and second-order drifting gratings were placed in various directions, and the tuning curves were analyzed using the modified circular variance \cite{mazurek2014robust}. The ME units responded primarily to first-order motions. In contrast, the higher-order channel was more sensitive to second-order motion. Sensitivity was further enhanced via training on non-diffuse materials (blue dots in Fig. \ref{fig-resultsec}-B).
		
		Additionally, we compared the Pearson correlations between model responses and motion GT across SOTA optical flow models, including RAFT, GMFlow, and multi-frame-based VideoFlow \cite{shi2023videoflow}. As shown in Fig. \ref{fig-resultsec}-A, our model exhibited the highest correlation and stability, closely matching human performance. Notably, although non-diffuse materials improved second-order motion perception in most models, current CV models still struggle with these motions, exhibiting lower correlations and higher variance. This limitation likely stems from structural differences. Our model, based on 3D spatiotemporal filters, emulates the separate human motion processing mechanisms of both first- and second-order perception. In contrast, CV models primarily track the absolute pixel correspondences between frames, and thus rely on pixel intensity \cite{lucas1981}. As second-order motions, such as drift-balanced motion, lack explicit pixel correspondences across frames, such models often become unstable and generate noise.
		\begin{figure*}[ht!]
			\begin{center}
				\includegraphics[width=1.0\textwidth]{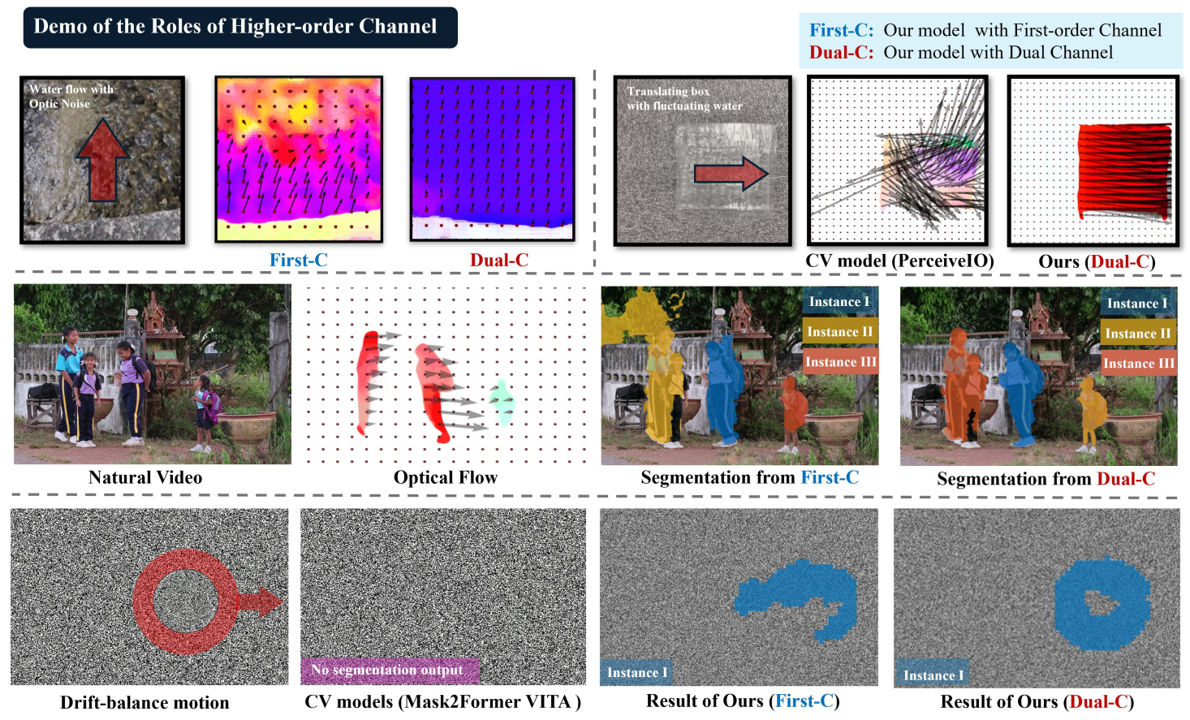}
			\end{center}
			\caption{{\bfseries \textit{Demonstration of the Roles Played by the Higher-order Channel}}{\bfseries }: We present several qualitative results derived when comparing the first-order and dual-channel approaches. First Row: Motion estimation in noisy natural conditions, compared with a state-of-the-art CV method \cite{jaegleperceiver}. Second Row: Instance segmentation based on motion features, where the dual-channel approach generally affords finer object segmentation. Third Row: The higher-order channel enhances the segmentation of drift-balanced motion. This remains undetectable by the SOTA CV segmentation methods \cite{heo2022vita}.
			}
			
			\label{f-firseccom}
		\end{figure*}
		\subsection{The Interplay Between the First- and Higher-order Channels}
		\label{sec2.4}
		Fig. \ref{f-firseccom} presents qualitative data illustrating the difference between the first- and second-order channels, demonstrating their function when processing natural scenes with noisy optical environments (First Row). Higher-order processing affords more stable results when interpreting global flow motion (Left). Such processing effectively tracks the movement of a plastic box with fluctuating water inside, even outperforming certain SOTA CV models \cite{jaegleperceiver} when handling such extremely noisy, but natural, scenes. The second and third rows show the segmentation results for both natural scenes and pure drift-balanced motion \cite{secondordermotion}. In terms of segmentation, the higher-order channel usually helps the model to identify objects in motion. The segmentation results are finer than those of the first-order channel alone. We verified these results using the DAVIS 2016 video segmentation benchmark \cite{DAVIS} (3,505 image samples). The dual-channel approach achieved a Mean Intersection over Union (IoU) score of $0.60$, thus higher than the single-channel score of $0.56$. In the last row, we show that our framework can identify objects based solely on motion, even when they are spatially invisible, as seen in the pure drift-balanced motion test. The higher-order channel affords a distinct advantage under such conditions, effectively identifying object instances within noise. Such second-order motion patterns are near-undetectable by current CV segmentation models, including SOTA video segmentation models \cite{cheng2022masked, heo2022vita}. Note that our segmentation results were obtained using a naive graph bipartition \cite{shi2000normalized} without additional training. For the technical details, please refer to Section \ref{sec4.13}.

	\section{Discussion}

		\label{sec3}
		We establish the first human-aligned optic flow estimation model sensitive to both first- and higher-order motions. The model successfully replicates the characteristics of human visual motion in various scenarios ranging from typical stimuli to more complex natural scenes, striking a good balance between human and machine vision models.
		
		Recent studies have also leveraged DNNs to infer the neural and perceptual mechanisms underlying visual motion. For example, Rideaux et al. \cite{rideaux2020but, rideaux2021exploring} and Nakamura and Gomi \cite{nakamura2023decoding} used multilayer feedforward networks, but Storrs et al. \cite{storrs2022properties} used a predictive coding network (PredNet), when building image-computable models of human visual motion processing. DorsalNet \cite{mineault2021your} employed first-person-perspective video stimuli to train a 3D ResNet model that predicted self-motion parameters and recapitulated the neural representation of the dorsal visual stream. Despite their contributions, these models cannot estimate the dense optical flows consistent with the physical or perceptual GTs, nor do they account for higher-order motion processing. 
		
		{\bfseries Modeling Visual Motion Processing:}
		We modeled human visual motion processing, including the V1-MT architecture, via motion-energy (ME) sensing and integration. Our key notions are that spatiotemporal tuning of the motion-sensing filter is trainable, and that motion integration can be modeled by a recurrent network based on the motion graph. After end-to-end training on a wide range of datasets, our model generalized both simple laboratory stimuli and complex natural scenes well. The model naturally captures various characteristics of neurons in the motion pathway, including the change in spatiotemporal tuning from the V1 to the MT areas. Motion integration successfully explains the physiological findings---the changes in the populations of component and pattern cells from the V1 to the MT---and also the psychophysical findings, adaptive global motion pooling. The utility of the attention mechanism during motion integration may be attributable to its similarity to the human visual grouping mechanism \cite{mehrani2023self} or to feature grouping/binding process including a top-down attentional selection mechanism \cite{tsotsos2005attending}.
		
		{\bfseries Second-order Motion Processing:}
		Another critical contribution of this study is that we reveal a functional role for second-motion perception. The ability to perceive the movement of higher-order features, such as contrast and temporal modulations, is a unique feature of biological visual motion processing. Second-order motion has received little attention from the CV community because the functional significance thereof has been poorly understood. Early studies suggested that visual analysis of second-order features might aid recognition of global image spatial structure \cite{daugman1995demodulation} and/or might distinguish separation by shading from a material change \cite{schofield2000does}. However, the significance of second-order motion remained unclear. Here, we show that biological systems may engage in second-order motion perception to ensure reliable estimation of the motions of naturally glossy or transparent materials that are rendered optically turbulent by reflection/refraction. This is an important advance in making CV algorithms more human-aligned and, at the same time, more robust in estimating the dynamic structural changes of natural scenes. Furthermore, our study shows that machine learning can afford conceptual proof of neuroscientific hypotheses that suggest how specific functions evolved in natural environments.
		
		{\bfseries Relationship with Computer Vision Models:}
		This study does not seek to outperform SOTA CV models that have been optimized for certain engineering tasks. Rather, we employ a heuristic approach to balance the alignment of human vision with robust processing of natural scenes. However, inspired by the human visual system, it may be possible to expand the capacities of CV models. For example, we show that human-aligned computation efficiently captures inherent human-perceived flow illusions that CV models often fail to replicate (Tab. \ref{t1}). The instability of current CV methods, when presented with certain scenarios, is usually because they seek to match the pixel correspondences between frame pairs \cite{ranjan2019attacking}. This strategy differs from the human motion perception mechanism, which depends on dynamic spatiotemporal features and demonstrates exceptional stability and adaptability in interpreting physical object motion. Furthermore, the second-order motion system extracts the long-range motions of high-level features well. The addition of this system not only combats noise and optical turbulence but also results in a more stable and reliable motion estimation model, particularly useful in challenging scenarios such as adversarial attacks \cite{ranjan2019attacking} or extreme weather conditions \cite{Schmalfuss2023Distracting}. We believe these advancements offer significant insights to enhance motion estimation in the CV field and develop a more reliable and stable model.

		{\bfseries Limitations:}
		Human-like visual systems require capabilities beyond basic motion energy computation, such as adaptive motion integration and higher-order motion feature extraction. Although our approach employs multilayer 3D CNNs and motion graphs to address these needs, this inevitably sacrifices some of the transparency and explainability compared to traditional human vision models. Interpreting the specific higher-order features being extracted remains challenging, as does understanding how dynamic graph structure might be implemented in actual neural systems. 
		These elements extend beyond motion perception and involve broader integrative functions of the human visual system \cite{nishida2018motion}. 
		
		Higher-order motion processing serves broader functions, including self-location and navigation in dynamic environments \cite{angelaki2005self,fencsik2007role,land1994we}, and hierarchical decomposition of motion and object inference \cite{gershman2016discovering, bill2022visual}. These aspects are not explicitly modeled here. However, our model exhibits grouping and segmentation capacities based on motion inference. These are important steps toward hierarchical inferences of natural scenes. We plan to explore these aspects further in an effort to advance human-like, higher-order motion processing.

	\section{Methods}\label{sec4}

		\subsection{Model Structure}
		\label{sec4.1}
		Our biologically oriented model features two stages, Stages I and II. As shown in Fig. \ref{fmodel}, Stage I has two channels, of which the first engages in straightforward luminance-based ME computation, and the second contains a multilayer 3D CNN block that enables higher-order feature extraction.
		\subsubsection{Stage I: First-order Channel}
		{\bfseries Spatiotemporally Separable Gabor Filter:}
		When building our image-computable model, each input was a sequence of grayscale images $\mathbf{S}(\mathbf{p},\mathbf{t})$ of spatial positions $\mathbf{p}=(x,y)$ within domain $\mathbf\Omega$ at times $\mathbf{t>0}$. We sought to capture local motion energies at specific spatiotemporal frequencies, as do the direction-selective neurons of the V1 cortex. We modeled neuron responses using 3D Gabor filters \cite{jones1987two,jones1987evaluation}. To enhance computational efficiency, these were decomposed into spatial 2D Gabor filters $\mathcal{G(\cdot)}$ and temporal 1D sinusoidal functions exhibiting exponential decay $\mathcal{T(\cdot)}$. Given the coordinates $x^{\prime}=x \cos \theta+y \sin \theta$ and $y^{\prime}=-x \sin \theta+y \cos \theta$, the filters may be defined as follows:

	\begin{equation}
		\left\{\small
		\begin{array}{lr}
			\mathcal{G}(x, y ; {\color{red}f_{s}}, {\color{red}\theta},{\color{red} \sigma}, {\color{red}\gamma})=\exp \left(-\frac{x^{\prime 2}+\gamma^2 y^{\prime 2}}{2 \sigma^2}\right) \cdot 
			e^{ \left(2 \pi {f_s}{x^{\prime}}\right)i }, & \\
			\mathcal{T}\left(t ; {\color{red} f_t},{\color{red}\tau}\right)=\exp{\left(-\frac{t}{\tau}\right)}\exp{( 2 \pi i \left(f_t t\right))},  &\\
			s.t.~ \{x,y,t~|~0 \le t < T; (x,y)~|(x^2 + y^2 \le R^2)\}
		\end{array}
		\right.
	\end{equation}

		Trainable parameters in red color include $f_s$, $f_t$, $\theta$, $\sigma$, and $\gamma$, controlling spatiotemporal tuning, orientation, and Gabor filter shape, while $\tau$ adjusts temporal impulse response decay. All parameters are subject to certain numerical constraints, such as $\theta$ being limited to [0,$2\pi$) to avoid redundancy; $f_s$ and $f_t$ are limited to less than 0.25 pixels per frame to avoid spectrum aliasing, etc.
		The response $L_n$ to the stimuli $\mathbf{S}(\mathbf{p},\mathbf{t})$ is computed via separate convolutions:

		\begin{align*}
			&L_n(x,y,t;\Theta)  = (\mathbf{S}*\mathcal{G})*\mathcal{T} \\
			& = \iiint \mathbf{S}(\mathcal{X}, \mathcal{Y}, \mathcal{T}) \cdot  \mathcal{G}_{n}(x-\mathcal{X}, y-\mathcal{Y}) \cdot \\
			& \mathcal{T}_{n}( t-\mathcal{T}) \, d\mathcal{X} \, d\mathcal{Y} \, d\mathcal{T} \quad +{\color{red}\alpha_1}
		\end{align*}where $\alpha_1$ are the learned spontaneous firing rates. 
		Further, local motion energy is captured by a phase-insensitive complex cell in the V1 cortex, which computes the squared summation of the response from a pair of simple V1 cells with orthogonal receptive fields \cite{spatiotemporal1985}, defined as (even and odd):

	\begin{equation}
		\left\{\small
		\begin{array}{lr}
			L_n^{o}(x,y,t;{\color{red}\Theta}) = \mathbf{S}*{\Im}[\mathcal{G}]*{\Re}[\mathcal{T}] +  \mathbf{S}*{\Im}[\mathcal{G}])*{\Im}[\mathcal{T}] & \\
			L_n^{e}(x,y,t;{\color{red}\Theta}) = \mathbf{S}*{\Re}[\mathcal{G})*{\Re}[\mathcal{T}] -  \mathbf{S}*{\Im}[\mathcal{G}]*{\Im}[\mathcal{T}],
		\end{array}
		\right.
	\end{equation}

		where ${\Re}(\cdot) $ and ${\Im}(\cdot)$ extract the real and imaginary parts of a complex number and $*$ denotes convolution operations. Then, the complex cell response $L_n^{c}$ is:
	\begin{equation}
		L_n^{c}(x,y,t;{\color{red}\Theta}) = 	\left(L_n^{o}(x,y,t;{\color{red}\Theta})\right)^2 + (L_n^{e}(x,y,t;{\color{red}\Theta}))^2
	\end{equation}

		{\bfseries Multiscale Wavelet Processing:}
		The convolution kernel of our spatial filter has a fixed size of $15\times15$. This imposes a physical limitation on the receptive field of each unit. We employed a multiscale processing strategy to enhance receptive field size flexibility. Specifically, we constructed a pyramid of eight images that were linearly scaled from $H\times W$ to $\frac{H\times W}{16}$. In total, 256 independent complex cells were deployed across different scales. This enabled representation of different groups of cells that were sensitive to short- and long-distance motions \cite{castet1999long}.	
		The N complex cells $\{L_n^c\}_{i}^{N}$ capture ME on multiple scales. We subjected each cell to energy normalization to ensure that the energy levels were consistent:

	\begin{equation} \small
		\hat{L}_n^c(t)=\frac{{{\color{red}K_1}} L_n^c(t)}{\sum_{\mathrm{i=1}}^{\mathrm N} L_{\mathrm{i}}^c(t)+{{\color{red}\sigma_1}}},
	\end{equation}

		where $\sigma_1$ is the semi-saturation constant of normalization and $K_1 > 0$ determines the maximum attainable response. We interpret the response, denoted $\hat{L}_n(t)$, as the model equivalent of a post-stimulus time histogram (PSTH), which is a measure of the neuron's firing rate. Physiologically, such responses could also be computed using inhibitory feedback mechanisms \cite{heeger1993modeling,carandini1994summation}. Bilinear interpolation was used to resize the multiscale motion energies to the same spatial size, thus $\frac{H \times W}{8}$. 
		In the DNN context, this balances the trade-off between the spatial resolution and the computational overhead. The final output of the first stage is a 256-channel feature map $\mathbf{E_1} \in \mathbb{R}^{\frac{H}{8}\times\frac{W}{8}\times256}$ that captures the underlying, local ME and thus partially characterizes the cellular patterns of the V1 cortex in a computational manner \cite{spatiotemporal1985}. 
		\subsubsection{Stage I: Higher-Order Channel}
		\label{sec4.12}
		In the second-order channel, we employ standard 3D CNNs to extract non-first-order features. This channel features five layers of 3D CNNs, each of kernel size \(3\times3\times3\), linked via residual connections and nonlinear ReLU activation functions. The 3D CNN layers engage in preprocessing before extraction of nonlinear features, which are then processed using the ME constraints described above, and the motion energies calculated. Each input to this channel is a sequence of RGB images, and the output is formatted to match that of the first-order channel, thus \(\mathbf{E_2} \in \mathbb{R}^{\frac{H}{8} \times \frac{W}{8} \times 256}\). Both the first- and second-order channel activations undergo the same normalization process, after which they are merged via a \(1 \times 1\) convolution. Each fused output, thus \(\mathbf{E_m} \in \mathbb{R}^{\frac{H}{8} \times \frac{W}{8} \times 256}\), is then fed to Stage II.
		In Figs. \ref{fig-resultsec} and \ref{f-firseccom}, we designate the model incorporating Stage II with \(\mathbf{E_m}\) as the "Dual Channel." Conversely, the model relying solely on \(\mathbf{E_1}\) is labeled as the "First-order Channel." The results presented in Figs. \ref{f-motionintegration} and \ref{f-motionintegration2} are exclusively derived from the model utilizing the First-order Channel.

		\subsubsection{ Stage II: Global Motion Integration and Segregation}
		\label{sec4.13}
		First-stage neurons have a limited receptive field, constraining them to detect only nearby motion. Solving the aperture problem in motion perception systems necessitates advanced spatial integration \cite{pack2001temporal}. This process involves complex mechanisms \cite{gilaie2016visual,noest1993role} and requires extensive prior knowledge, which may surpass traditional modeling methods. CNNs, with their extensive parameterization and adaptability, provide a viable solution. However, spatial integration of local motions demands more versatile connectivity than that offered by standard convolutions. Such convolutions are limited to local receptive fields \cite{wei2018revisiting}. To address this, we developed a computational model that employed a graph network and recurrent processing for effective motion integration.
		
		\textbf{Motion Graph Based on a Self-Attention Mechanism:} We move beyond traditional Euclidean space in images, creating a more flexible connection across neurons using an undirected weighted graph, \(\mathbf{G} = \{\mathbf{V}, \mathbf{A}\}\). Here, \(\mathbf{V}\) denotes nodes (each spatial location \(p(i,j)\)) and \(\mathbf{A}\) is the adjacency matrix, indicating connections among nodes.  The feature of each node is the entire set of the corresponding local motion energies, thus \(\mathbf{E}(i,j) \in \mathbb{R}^{1 \times 256}\). The connection between any pair of nodes is computed using a specific distance metric. Strong connections form between nodes with similar local ME patterns. This allows the model to establish connections flexibly between different moving objects or elements across spatial locations, thus creating what we term a motion graph.
		Specifically, the distance between any pair of nodes $(i,j)$ is calculated using the cosine similarity. This is similar to the self-attention mechanisms of current transformer structures \cite{attention2017,wang2018non,dosovitskiy2020image}.
		We use the adjacency matrix $\mathbf{A} \in \mathbb{R}^{HW\times HW}$ to represent the connectivity of the whole topological space, where $\mathbf{A}$ is a symmetrical, semi-positive definite matrix defined as:

	\begin{equation}
		\mathbf{A}(i,j) = 	\mathbf{A}(j,i) =  \frac{\varphi(\mathbf{E})_i \cdot \varphi(\mathbf{E})_j}{\|\varphi(\mathbf{E})_i\|\|\varphi(\mathbf{E})_j\|} .
	\end{equation}

		We subject the connections between graphs to exponential scaling using the matrix $\mathbf{A}$ given by $\exp(\mathbf{A}{\color{red}s})$, where ${\color{red}s}$ is a learnable scalar restricted to within (0,10) to avoid overflow. The smaller the $s$, the smoother the connections across nodes, and vice versa. Finally, a symmetrical normalization operation balances the energy, thus $\mathbf{A} := \mathbf{D}^{-\frac{1}{2}} \exp({\color{red}s}\mathbf{A})\mathbf{D}^{-\frac{1}{2}} $, where $\mathbf{D}$ is the degree matrix.
		This yields an energy-normalized undirected graph. Intuitively, the adjacency matrix represents the affinity or connectivity of a neuron within the space. Strong global connections form between neurons, the motion responses of which are related.

		\textbf{Recurrent Integration Processing:} Recurrent neural networks flexibly model temporal dependencies and feedback loops, which are fundamental aspects of neural processing in the brain \cite{serre2019deep}. We use a recurrent network, rather than multiple feedforward blocks, to simulate the process of local motion signals being gradually integrated into the MT and eventually converge to a stable state.
		
		During each iteration (i), an adjacency matrix $\mathbf{A^i}$ is first constructed using the current graph embedding feature $A^i$. Subsequent motion integration is achieved through a simple matrix multiplication. The integrated motion information is then passed through convolutional GRU blocks that update the ME:

	\begin{equation}
		\mathbf{E}^{i+1} = \operatorname{GRU}_{\color{red}\theta}(\mathbf{A}^i  \times \mathbf{E}^i, \mathbf{E}^i)
	\end{equation}

		This is computationally similar to the information propagation mechanisms of transformers \cite{attention2017,wang2018non} and could also be viewed as a simplified form of graph convolution \cite{kipf2016semi}. 
		The GRU is a gated recurrent unit \cite{chung2014empirical} implemented in a convolutional manner. We adopt a separable spatiotemporal approach that is guided by RAFT \cite{raft}. Such motion integration approximates the ideal, final convergence state of ME $\mathbf{E}_k \rightarrow \mathbf{E}^*$. This is achieved via recurrent iteration.

		We adopted a same approach to decode the 2D optical flow from  $\mathbf{E}$ of each iteration $k$. 
		Specifically, the integrated motion $\mathbf{E}$ is squared to ensure positivity and then normalized in terms of energy:

	$$\hat{\mathbf {E}}(i,j)={\small {\color{red}K_2} \mathbf {E}^2(i,j)} / {\small \sum_{{i,j}}^{\mathrm H \mathrm W} \mathbf{E}^2(i,j)+{\color{red}\sigma_2}^2}.$$

		This yields $\hat{\mathbf {E}}\in \mathbb{R}^{H\times W \times 256}$, which could be viewed as a post-stimulus time histogram of neuronal activation. We use a shared flow decoder to project the activation pattern of each spatial location onto the motion field $F \in \mathbb{R}^{H\times W \times 2}$. This decoder employs multiple $1\times1$ convolution blocks with residual connections, as do recent advanced optical flow models \cite{pwcnet,arflow}. We observed that the results generally converged by the fourth iteration. This was therefore chosen as the standard Stage II output.

		\textbf{Cutting of an Object Instance from the Motion Graph:}
		The interactions of objects in a dynamic scene are reflected in the adjacency matrix of the motion graph \( \mathbf{G} \). After incorporation of this adjacency matrix into $\mathbf{A} \in \mathbb{R}^{HW\times HW}$, segmentation can be achieved using a graph-cut method. Specifically, we employ the Normalized Cuts (Ncut) method \cite{shi2000normalized}. This partitions a graph into disjoint subsets by minimizing the total edge weight between the subsets relative to the total edge weight within each subset.
		Specifically, the Laplacian matrix of \( \mathbf{G} \) can be expressed as \( \mathbf{L} = \mathbf{D} - \mathbf{A} \), or, in the symmetrically normalized form, as:  
		\(
		\mathbf{L} = \mathbf{I_n} - \mathbf{D}^{-\frac{1}{2}}\mathbf{A}\mathbf{D}^{-\frac{1}{2}},
		\)
		where \( \mathbf{D} \) is a diagonal matrix defined as \( \mathbf{D} = \text{diag}\left(\left\{\sum_j A_{ij}\right\}_{j=1}^n\right) \). \( L \) is a semi-positive definite matrix, which facilitates the orthogonal decomposition that yields:  
		\(
		\mathbf{L} =\mathbf{U}\Lambda \mathbf{U}^T,
		\)
		where \( \mathbf{U} \) is the set of all orthonormal basis vectors, denoted as \( \{u_i\}_{i=1}^n \), thus the Fourier basis of \( \mathbf{G} \). \( \Lambda \) is a diagonal matrix containing all eigenvalues \( \{\lambda_i\}_{i=1}^n \) ordered as \( \lambda_1 \leq \lambda_2 \leq \dots \leq \lambda_n \). According to \cite{shi2000normalized}, the eigenvector corresponding to the second smallest eigenvalue, \( u_2 \in \mathbb{R}^{HW}\), commonly termed the Fiedler vector, yields a real-valued solution to the relaxed Ncut problem. In our implementation, we extract \( u_2 \) and then apply binarization using the rule \( u_2 = u_2 > \text{mean}(u_2) \). The resulting binary segmentation is viewed as a potential field and further refined using a conditional random field \cite{chen2017deeplab}. As such binarization does not inherently distinguish between foreground and background, we adaptively assign a polarity that matches the foreground during evaluation using the DAVIS 2016 segmentation benchmark. The results shown in the second row of Fig. \ref{f-firseccom} were obtained using a recurrent bipartitioning method \cite{wang2023cut} that allows multi-object segmentation. Notably, the entire process is training-free.
		\subsection{Training Strategy}
		\label{sec4.2}
		We employ a supervised learning approach guided by the similarity between human motion perception and the physical GT, as suggested by \cite{Yang2023}. However, our primary focus is on how effectively the model mimics human motion perception, rather than how precisely it predicts the GT. During training, we utilize a sequential pixel-wise mean-square-error loss to minimize the difference between the GT and the model predictions of Stage I and of each iteration of Stage II.
		\subsubsection{Dataset}
		Our dataset includes a variety of natural- and artificial-motion scenes. Specifically, it incorporates existing benchmarks such as MPI-Sintel and Sintel-Slow \cite{sintel}, as well as natural videos from DAVIS, with pseudo-labels generated using FlowFormer \cite{flowformer}. This collection is termed Dataset A.
		Additionally, we included custom multi-frame datasets with simple non-textured 2D motion patterns (Dataset B) and drifting grated motions (Dataset C). These datasets introduce basic motion patterns that train the model from scratch, accelerating convergence and enhancing stability. They also aid model adaptation to non-textured scenarios and introduce a slow-world Bayesian prior \cite{weiss2002motion} that addresses ambiguous 1D motions.
		To study second-order motion, we developed datasets with diffuse (Dataset D) and non-diffuse (Dataset E) objects and integrated them into training. We then evaluated how the model perceived material properties and second-order motion. We define three training types:
		\begin{enumerate}
			\item \textbf{Type-I and Type-II:} The model was trained separately on D and E to assess material properties related to second-order motion perception. The results are shown in Fig. \ref{fig-resultsec}-B and -C).
			\item \textbf{Type-III:} The model was trained on a mixed dataset \{A, B, C, D, E\} using a curriculum strategy, thus starting with \{B, C\} and progressing to the full set. This approach, commonly used during optical flow model training \cite{flownet2, raft}, improves convergence and robustness. All other results are based on Type-III training.
		\end{enumerate}
		\subsubsection{ The Environment:} 
		We performed model training using the PyTorch 2.0 framework on a workstation with Five Nvidia RTX A6000 GPUs operating in parallel with the CUDA 11.7 runtime. Data analysis employed MATLAB 2021 and Python libraries, including Numpy, SciPy, and Pandas.
		\subsubsection{Timing:} 
		Given the standard playback frame rate of 25 frames per second and the human visual impulse response duration of approximately 200 ms, we configured the temporal window of Stage I to cover six frames (200 ms). For the first-order channel, sequences of 11 consecutive grayscale images were input. Supervised training uses the instantaneous velocity at the sequence midpoint, thus frame 5th, as the training label. The higher-order channel with the 3D-CNN was trained using a longer temporal sequence of 15 frames to capture long-term spatiotemporal features effectively.
		\subsection{Dataset Generation}
		\subsubsection{Dataset Rendering}
		We used the Kubric pipeline \cite{greff2021kubric} to generate large motion datasets that integrated Py-bullet \cite{Coumans2016Pybullet} for physics simulation and Py-blender \cite{Blender2021} for rendering. We selected several 3D models and textures from ShapeNet and GSO, and natural HDRI backgrounds from Polyhaven \cite{Zaal2021Polyhaven}. The latter are realistic in terms of lighting and texture.
		Objects were positioned at specific heights in 3D scenes. We used Py-bullet to simulate physical dropping and interactions between objects. The camera placements were carefully chosen to optimize the viewing angles. Blender then rendered these 3D scenes into 2D images, simultaneously generating a detailed motion flow field for each pixel, as shown in Fig. \ref{fsecdata}.
		Material properties were manipulated using the \verb|principled BSDF| function to create apparently natural optical effects. We generated materials with both Lambertian and non-Lambertian reflections that differed in terms of metallicity, specularity, anisotropy, and transmission. All other features, thus the optical flow distribution, illumination, object, and scene arrangements, were standardized to maintain consistency across the dataset.
		\subsubsection{Second-order Dataset Modulation}
		\label{sec4.32}
		We developed a second-order dataset to benchmark second-order motion perception in both humans and computational models. The dataset consists of 40 scenes featuring seven types of second-order motion modulations. Each modulation comprises 16 frames, with a randomly moving carrier overlaid on a natural image background. To eliminate first-order motion interference, the natural images were kept static, and the random motion patterns were generated using a Markov chain. The random carrier motion, \( \mathbf{S}(t) = [U(t), V(t)] \), evolved according to the transition probability:

	\[
	\operatorname{Pr}[\mathbf{S}(t)] = \operatorname{Pr}[\mathbf{S}(t) = s_t \mid \mathbf{S}(t-1) = s_{t-1}],
	\]

		where the motion states \( [U, V] \) were sampled from 2D Gaussian distributions conditioned on the previous state. The carrier was subjected to seven distinct second-order motion modulations, encompassing spatial effects, such as \{Gaussian blurring\}; temporal effects, such as \{drift-balanced\} motion and \{shuffle Fourier phase\}; and spatiotemporal effects like \{water waves\} and \{swirls\}. The spatial noise and blur were sparse Gaussian noise and localized Gaussian blur, respectively. The water wave, swirl, and random flow field modulations warp pixels using specific flow fields. In terms of the water wave dynamics, the flow field \( f_{u, v, t} = \left[\frac{\partial K}{\partial x}, \frac{\partial K}{\partial y}\right] \) was:

	\[
	\begin{aligned}
		K(r, t) &= \cos(2 \pi f r) \cdot e^{-\gamma r^2} \cdot \cos(2 \pi \xi t) \cdot e^{-\delta t^2}, \\
		r &= \sqrt{x^2 + y^2},
	\end{aligned}
	\]

		where \( f \), \( \xi \), and \( \delta \) control the wave frequency, temporal variation, and damping respectively. We superimposed multiple water waves that differed in terms of their dynamics in different locations. This created chaotic, local optical turbulence contemporaneous with carrier motion. The real carrier motion was thus obscured by local optical noise and was invisible in Fourier space, epitomizing the characteristics of second-order motion. Similarly, \{random flow field\} or \{shuffle Fourier phase\} modulation involves warping of either the pixels or the Fourier phase of original local regions using a randomly sampled Gaussian flow field.
		\subsection{Experimental Details}
		\label{sec4.4}
		\subsubsection{In-silico Neurophysiological Methods}
		
		We employed drifting Gabor or plaid (composed of two Gabor components) with a single frequency component as the input stimulus. For second-order motion, drift-balanced motion modulation was applied to the same Gabor envelope.
		
		The model responses after Stage I and after each iteration of Stage II were considered analogous to the PSTH of a neuron, thus reflecting activation levels. Responses across the spatial dimensions were averaged to obtain the activation distributions of the 256 units, represented as $\mathbb{R}^{1\times 1 \times 256}$ with respect to the input stimulus. The stimuli were typically $512\times512$ pixels in size, with full contrast.
		
		\textbf{Directional Tuning:} We employed a single frequency drifting-Gabor and a plaid (superimposed at $\pm 30$ degrees) as stimuli. Initially, 12 directions were uniformly sampled from $(0,2\pi]$. For each direction, we logarithmically sampled $8 \times 8 = 64$ sets of spatiotemporal frequency combinations and used the drifting-Gabor stimulus to obtain 64 directional tuning curves for each unit. The spatiotemporal frequency with the largest s.t.d. was selected as the preferred frequency $st^*$ for each unit. Subsequently, Gabor and plaid stimuli with the frequency configurations of $st^*$ were input to the model to derive the directional tuning curves of all units. The model tuning curve with $st^*$ as the drifting-Gabor was termed $\mathcal{C}$ and that for the plaid $\mathcal{P}$. We next assessed the directional tuning capacity by deriving partial correlations \cite{movshon1992analysis}:

	\begin{equation}
		\left\{
		\begin{array}{l}
			R_{\text{pattern}} = \frac{r_p - r_c r_{cp}}{\sqrt{(1 - r_c^2)(1 - r_{cp}^2)}}, \\
			R_{\text{component}} = \frac{r_c - r_p r_{cp}}{\sqrt{(1 - r_p^2)(1 - r_{cp}^2)}},
		\end{array}
		\right.
	\end{equation}

		where $r_c$ is the correlation between $\mathcal{P}$ and the component prediction that is the superimposed $\pm 30$-degree shift of $\mathcal{C}$. $r_p$ is the correlation between $\mathcal{P}$ and the pattern prediction $\mathcal{C}$). $r_{cp}$ is the correlation between these two predictions. Units were classified as ``component", ``pattern", or ``unclassified" based on these correlations, as illustrated in Fig. \ref{f-motionintegration}-A.
		
		\textbf{Orientation Selectivity Quantification:} Fig. \ref{fig-resultsec}-B shows how the orientation selectivity $O_{\text{ori}}$ was quantified using the modified circular variance \cite{mazurek2014robust}:

	\begin{equation}
		\small
		O_{\text{ori}} = \left|\frac{\sum_i A(\theta_i) \exp(2 i \theta_i)}{\sum_i A(\theta_i)}\right|,
	\end{equation}

		where $A(\theta_i)$ is the normalized response at angle $\theta_i$.
		\subsubsection{Human Comparison}
		\label{sec4.4.2}
		We used the human-perceived flow data \cite{Yang2023} of the Sintel benchmark for comparison. The metrics employed included the vector endpoint error, the Pearson correlation, and a partial correlation. As each model used for comparison had been trained to estimate GT, the partial correlation was crucial. This measures the relationship between human responses and model predictions after controlling for the GT:

	\begin{equation} \small
		r_{\text{resp model} \cdot GT} = \frac{r_{\text{resp model}} - r_{\text{resp} GT} \cdot r_{\text{model} GT}}{\sqrt{1 - r_{\text{resp} GT}^2} \sqrt{1 - r_{\text{model} GT}^2}},
	\end{equation}
	
		where $r$ is the Pearson correlation. Additionally, we collected human data using our second-order benchmark.

		\textbf{Participants and Apparatus:} Stimuli were displayed on a VIEWPixx /3D LCD monitor (VPixx Technologies) with a resolution of $1920 \times 1080$ pixels at a 30-Hz refresh rate. The display luminance levels were linearly calibrated using an i1Pro chromometer (VPixx Technologies). The minimum, mean, and maximum values were 1.8, 48.4, and $96.7 cd/m^2$ respectively. The viewing distance was 70 cm and each pixel subtended 1.2376 arcmin.
		Participants sat in a darkened room using a chinrest to stabilize the head and performed experiments from the Psychopy library \cite{Peirce2019PsychoPy2}. The study adhered to the ethical standards of the Declaration of Helsinki, with the exception of preregistration, and was approved by the Ethics Committee of Kyoto University (approval no. KUIS-EAR-2020-003). Two authors and one naive participant (three males, average age 25.3 years) with normal or corrected-to-normal vision participated. Informed consent was obtained prior to the experiment. All participants were later financially compensated.

		\textbf{The Stimuli and the Procedure:}
		Following \cite{Yang2023}, the protocol evaluated the ability of the participants to estimate second-order motion. Each trial displayed target movement within apertures 600 pixels in diameter at the center of the screen for 500 ms (15 frames), followed by brown noise movements 120 pixels in diameter of the same duration with interstimulus intervals of 750 ms. A probe 15 pixels in diameter was used to mark the location and timing of each target motion. Four dots, each of five pixels, served as position markers. The dots were orthogonally arranged 60 pixels distant from the display center. Each participant used a mouse to adjust the direction and speed of noise movement to that of the labeled motion of the target. Seven types of second-order motion modulations were tested, each over 40 scenes. To counter any directional bias, four variations were presented: original, horizontal flip, vertical flip, and both. In total, 1,120 trials per participant were conducted over 6 hours. The results were averaged across the flips to yield 280 perceived-motion vectors that were compared to those of CV models. The stimulus sequence was fully randomized for each participant.

	\backmatter

	\bmhead{Acknowledgments}
	
	This work was supported in part by JST, through the establishment of university fellowships for the creation of science and technology innovation, Grant Number JPMJFS2123; and in part by JSPS Grants-in-Aid for Scientific Research (KAKENHI), Grant Numbers JP20H00603, JP20H05957, and 24H00721. We thank Kubric \cite{greff2021kubric} for providing their data generation pipeline.
	
	\bmhead{Code Availability}
	Our model implementation and human experimental code are available at: \url{https://github.com/anoymized/multi-order-motion-model}.  
	The project website, featuring informative video demos, is available at: \url{https://anoymized.github.io/motion-model-website/}.

	\bibliographystyle{sn-nature} 
	\bibliography{nips}

\end{document}